\documentclass[conference]{IEEEtran}

\usepackage{cite}
\usepackage{bm}
\usepackage{amsmath}
\usepackage{bbm}
\usepackage{algorithmic}
\usepackage{graphicx}
\usepackage{textcomp}
\usepackage{xcolor}
\def\BibTeX{{\rm B\kern-.05em{\sc i\kern-.025em b}\kern-.08em
    T\kern-.1667em\lower.7ex\hbox{E}\kern-.125emX}}

\usepackage[T1]{fontenc} 
\usepackage[utf8]{inputenc}
\usepackage[american]{babel}
\usepackage[hidelinks]{hyperref}
\usepackage[capitalize]{cleveref}
\usepackage{mathtools}
\usepackage{booktabs}
\usepackage{fancyhdr}
\usepackage[shortcuts]{extdash}
\usepackage[all]{foreign}
\usepackage{bm}
\usepackage{tikz}
\usetikzlibrary{bayesnet}
\usepackage{siunitx}
\usepackage{xfrac}
\usepackage{amsfonts}
\usepackage{booktabs}
\usepackage{arydshln}
\usepackage{comment}
\usepackage{subcaption}
\usepackage[helvratio=0.92]{newtxtext}
\usepackage{temporal-logic}
\usepackage{amsthm}
\usepackage{amssymb}  

\usepackage[framemethod=tikz]{mdframed}

\usepackage{glossaries}

\newacronym{ad}{AD}{Annotated Disjunction}
\newacronym{ann}{ANN}{Artificial Neural Network}
\newacronym{ap}{AP}{Access Point}
\newacronym{cfr}{CFR}{Channel Frequency Response}
\newacronym{cmis}{CMISymb}{Conditional Mutual Information on Symbolic data}
\newacronym{cnn}{CNN}{Convolutional Neural Network}
\newacronym{dnn}{DNN}{Deep Neural Network}
\newacronym{csi}{CSI}{Channel State Information}
\newacronym{cv}{CV}{Computer Vision}
\newacronym{dl}{DL}{Deep Learning}
\newacronym{edl}{EDL}{Evidential Deep Learning}
\newacronym{har}{HAR}{Human Activity Recognition}
\newacronym{char}{CHAR}{CSI-based HAR}
\newacronym{charl-tre}{CHARL-TRE}{CHAR Latent Temporal Rule Extraction}
\newacronym{kl}{\mbox{KL}}{Kullback–Leibler}
\newacronym{lan}{LAN}{Local-Area Network}
\newacronym{lif}{LIF}{Leaky Integrate-and-Fire}
\newacronym{lpcmci}{LPCMCI}{Latent PCMCI}
\newacronym{lstm}{LSTM}{Long Short-Term Memory}
\newacronym{ltl}{LTL}{Linear Temporal Logic}
\newacronym{mimo}{MIMO}{Multiple-Input Multiple-Output}
\newacronym{mlp}{MLP}{Multi-Layer Perceptron}
\newacronym{mse}{MSE}{Mean Squared Error}
\newacronym{nad}{nAD}{Neural Annotated Disjunction}
\newacronym{nic}{NIC}{Network Interface Card}
\newacronym{nn}{NN}{Neural Network}
\newacronym{ood}{OoD}{Out-of-Distribution}
\newacronym{ofdm}{OFDM}{Orthogonal Frequency-Division Multiplexing}
\newacronym{ofdma}{OFDMA}{Orthogonal Frequency-Division Multiple Access}
\newacronym{pcmci}{PCMCI}{Peter-Clark Momentary Conditional Independence}
\newacronym{phy}{PHY}{Physical Layer}
\newacronym{sdd}{SDD}{Sentential Decision Diagrams}
\newacronym{sdr}{SDR}{Software-Defined Radio}
\newacronym{siso}{SISO}{Single-Input Single-Output}
\newacronym{snn}{SNN}{Spiking Neural Network}
\newacronym{std}{STD}{Standard Deviation}
\newacronym{stdp}{STDP}{Spike-Timing-Dependent Plasticity}
\newacronym{sta}{STA}{station}
\newacronym{vae}{VAE}{Variational Auto-Encoder}
\newacronym{wmc}{WMC}{Weighted Model Counting}
\newacronym{wlan}{WLAN}{wireless Local-Area Network}
\newacronym{gelu}{GELU}{Gaussian Error Linear Unit}
\newcommand*\wifi{\mbox{Wi-Fi}\xspace}
\newtheorem{definition}{Definition}

\begin{document}

\title{Towards Causally Interpretable \wifi CSI-Based Human Activity Recognition with Discrete Latent Compression and LTL Rule Extraction}

\author{
    \IEEEauthorblockN{
        Luca Cotti\IEEEauthorrefmark{1},
        Luca Lavazza\IEEEauthorrefmark{1}\IEEEauthorrefmark{2},
        Marco Cominelli\IEEEauthorrefmark{3},
        Liying Han\IEEEauthorrefmark{5},
        Gaofeng Dong\IEEEauthorrefmark{5},
        Francesco Gringoli\IEEEauthorrefmark{1},\\
        Mani B. Srivastava\IEEEauthorrefmark{5},
        Trevor Bihl\IEEEauthorrefmark{6},
        Erik P. Blasch\IEEEauthorrefmark{7},
        Daniel O. Brigham\IEEEauthorrefmark{7},
        Kara Combs\IEEEauthorrefmark{7},\\
        Lance M. Kaplan\IEEEauthorrefmark{4},
        and Federico Cerutti\IEEEauthorrefmark{1}\IEEEauthorrefmark{8}\IEEEauthorrefmark{9}\IEEEauthorrefmark{10}
    }\\
    \IEEEauthorblockA{
        \IEEEauthorrefmark{1}%
        University of Brescia, Italy.\
        \{luca.cotti, luca.lavazza, francesco.gringoli, federico.cerutti\}@unibs.it
    }
    \IEEEauthorblockA{
        \IEEEauthorrefmark{2}%
        Università La Sapienza Roma, Italy.\
    }
    \IEEEauthorblockA{
        \IEEEauthorrefmark{3}%
        DEIB,
        Politecnico di Milano, Italy.
        marco.cominelli@polimi.it
    }
    \IEEEauthorblockA{
        \IEEEauthorrefmark{5}%
        ECE Department,
        University of California, Los Angeles, USA.\
        \{liying98, gfdong, mbs\}@ucla.edu
    }
    \IEEEauthorblockA{
        \IEEEauthorrefmark{6}%
        Ohio University, USA.\
        bihlt@ohio.edu
    }
    \IEEEauthorblockA{
        \IEEEauthorrefmark{7}%
        Air Force Research Laboratory, USA.\
        \{erik.blasch.1, daniel.brigham, kara.combs.1\}@us.af.mil
    }
    \IEEEauthorblockA{
        \IEEEauthorrefmark{4}%
        U.S. Army DEVCOM Army Research Laboratory, USA.\
        lance.m.kaplan.civ@army.mil
    }
    \IEEEauthorblockA{
        \IEEEauthorrefmark{8}%
        Cardiff University, UK.\
    }
    \IEEEauthorblockA{
        \IEEEauthorrefmark{9}%
        University of Southampton, UK.\
    }
    \IEEEauthorblockA{
        \IEEEauthorrefmark{10}%
        Imperial College London, UK.\
    }
}

\maketitle

\begin{abstract}
    We address \gls{har} utilizing \wifi \gls{csi} under the joint requirements of causal interpretability, symbolic controllability, and direct operation on high-dimensional raw signals. Deep neural models achieve strong predictive performance on \gls{char}, yet rely on continuous latent representations that are opaque and difficult to modify; purely symbolic approaches, in contrast, cannot process raw \gls{csi} streams. We propose a fully automatic and strictly decoupled pipeline in which \gls{csi} magnitude windows are compressed by a categorical variational autoencoder with Gumbel-Softmax latent variables under a capacity-controlled objective, yielding a compact discrete representation. The encoder is then frozen and used as a deterministic mapping to one-hot latent trajectories. Causal discovery is performed on these trajectories to estimate class-conditional temporal dependency graphs. Statistically supported lagged dependencies are translated into \gls{ltl} rules, producing a fully symbolic and deterministic classifier based solely on rule evaluation and aggregation, without any learned discriminative head. Because rules are defined over discrete latent variables, antenna-specific rule sets can in principle be combined at the symbolic level, enabling structured multi-antenna fusion without retraining the encoder. Results from \gls{charl-tre} indicate competitive performance while preserving explicit temporal and causal structure, showing that deterministic symbolic classification grounded in unsupervised discrete latent representations constitutes a viable alternative to end-to-end black-box models for wireless \gls{har}.
\end{abstract}


\section{Introduction}\label{sec:introduction}
\wifi \gls{char} enables device-free sensing by exploiting motion-induced perturbations of wireless channels~\cite{wangUnderstandingModelingWiFi2015,zhangWidar30ZeroeffortCrossdomain2022}. The signal is however high-dimensional, noisy, and temporally structured. Consequently, most high-performing approaches rely on \glspl{dnn} that learn representations and decision boundaries directly from data~\cite{wangUnderstandingModelingWiFi2015}. While effective, such \gls{dnn} models provide limited interpretability: internal states are continuous, distributed, and not directly amenable to inspection or modification.

Our previous works addressed this tension between performance and interpretability from complementary perspectives~\cite{cominelliAccuratePassiveRadar2023,brescianiPreliminaryInsightsResourceconstrained2025}. In~\cite{cominelliAccuratePassiveRadar2023}, we investigated generative compression of multi-antenna \gls{csi} using \glspl{vae}, focusing on representation learning and signal fusion strategies. In~\cite{brescianiPreliminaryInsightsResourceconstrained2025}, we proposed a neuro-symbolic framework in which causal discovery was applied to externally extracted symbolic activities derived from video, and the resulting temporal logic constraints were integrated with neural processing of \gls{csi}. Although these approaches demonstrated the feasibility of combining wireless sensing with causal and logical reasoning, they either entangled compression and downstream reasoning or relied on externally provided semantic primitives.

\gls{charl-tre}\footnote{\url{https://github.com/LucaCtt/charl-tre}} addresses the problem of combining (i) raw \gls{csi} processing, (ii) causally interpretable temporal structure, and (iii) deterministic classification in a formal logical language, while explicitly improving upon our prior formulations. We propose a fully automatic pipeline that strictly decouples representation learning from symbolic reasoning and eliminates any reliance on externally defined features or human intervention. Raw \gls{csi} windows are compressed into a compact discrete latent representation using a categorical variational autoencoder with Gumbel-Softmax latent variables~\cite{kingmaAutoEncodingVariationalBayes2022,jangCategoricalReparameterizationGumbelsoftmax2017,maddisonConcreteDistributionContinuous2017}. After training, the encoder is frozen and is used as a deterministic mapping to hard one-hot latent codes, yielding a binary latent time series. Causal discovery is performed in this learned latent space using \gls{lpcmci}~\cite{gerhardusHighrecallCausalDiscovery2020} with ParCorr tests~\cite{rungeDetectingQuantifyingCausal2019}.

Unlike~\cite{brescianiPreliminaryInsightsResourceconstrained2025}, no semantic labels or video-derived symbolic activities are introduced: the latent space is learned exclusively from \gls{csi}, and the causal structure is inferred directly from the compressed wireless representation and translated into \gls{ltl} rules~\cite{pnueliTemporalLogicPrograms1977,baierPrinciplesModelChecking2008}. These rules form a symbolic, deterministic classifier applied after compression.

Differently from hybrid neuro-symbolic systems in which neural and symbolic components are interwoven~\cite{manhaeveDeepProbLogNeuralProbabilistic2018}, the present \gls{lpcmci} approach confines the neural component strictly to unsupervised compression and performs reasoning only in the discrete latent domain. The contribution is a modular framework for deriving a deterministic \gls{ltl} classifier from statistically inferred temporal structure in a learned categorical latent space. The resulting system maintains causal interpretability and enables rule-level modifications without retraining the encoder. Additionally, due to its discrete, rule-based structure, \gls{lpcmci} supports multi-antenna fusion by logically combining antenna-specific rule sets, without modifying the learned encoder. This system can also operate directly on high-dimensional wireless sensing data without requiring external feature engineering.

\section{Background}\label{sec:background}
\subsection{Activity Recognition Approaches}\label{sec:harapproaches}

There are two primary approaches to activity recognition: \textit{declarative} and \textit{data-driven}. Declarative approaches~\cite{storfRulebasedActivityRecognition2009,theekakulRulebasedApproachActivity2011,atzmuellerExplicativeHumanActivity2018} provide classification rules that can be utilized to define the activity. An example of such a rule in natural language could be: \textit{running means going steadily by springing steps so that both feet leave the ground for an instant in each step}~\cite{merriam-websterRun2026}. However, the types of input data declarative approaches can handle are often limited. Specifically, declarative rules typically require direct processing of the input data, which can pose challenges for unstructured data such as \wifi \gls{csi} (further discussed in \Cref{ssec:dataset}). Indeed, the authors are unaware of any declarative approaches for \gls{har} operating over \gls{csi} data.

Data-driven approaches (\eg~\cite{meneghelloSHARPEnvironmentPerson2023,bahadoriReWiSReliableWifi2022,liuHumanOccupancyDetection2020,cominelliAccuratePassiveRadar2023}) are specifically designed to handle data types for which it is difficult to define rules directly. Despite their advantages, data-driven approaches are more opaque and less flexible than their declarative counterparts, often making it impossible for the system's end-user to define patterns entirely. Several \gls{har} systems work by deriving some physically-related quantity from some sensors (\eg the \glspl{csi}), which is then used to train a deep learning classification system~\cite{meneghelloSHARPEnvironmentPerson2023,bahadoriReWiSReliableWifi2022,liuHumanOccupancyDetection2020}.

\subsection{Categorical Variational Autoencoders with Gumbel-Softmax}

\glspl{vae}~\cite{kingmaAutoEncodingVariationalBayes2022} are generative latent-variable models that learn a probabilistic encoder-decoder pair by maximizing a variational lower bound on the data likelihood. Given an observation $\bm{x}$ and latent variables $\bm{z}$, the objective balances a reconstruction term and a Kullback-Leibler (KL) divergence regularizer between the approximate posterior $q_{\phi}(\bm{z}\mid\bm{x})$ and a prior $p(\bm{z})$.

Standard \glspl{vae} assume continuous latent variables, typically Gaussian. However, discrete latent variables are often preferable when a symbolic interpretation or downstream logical reasoning is required. Direct optimization with categorical variables is hindered by the non-differentiability of sampling. The Gumbel-Softmax (or Concrete) distribution~\cite{jangCategoricalReparameterizationGumbelsoftmax2017,maddisonConcreteDistributionContinuous2017} provides a continuous relaxation of categorical sampling, enabling low-variance gradient-based optimization through the reparameterization trick.

In this work, we employ a categorical \gls{vae} in which the latent space comprises multiple independent categorical variables. After training, the encoder is frozen and hard one-hot codes are obtained via \textit{argmax}. This encoder produces a binary latent representation that is amenable to symbolic post-processing. Importantly, while the learned latent dimensions do not carry predefined semantics, their statistical and temporal regularities can be analyzed and manipulated at the symbolic level.

\subsection{Dataset}\label{ssec:dataset}

We rely on a \gls{csi} dataset\footnote{\url{https://github.com/ansresearch/exposing-the-csi}} publicly released by members of the author list. Full details are available in~\cite{cominelliExposingCSISystematic2023}. Briefly, the CSI dataset includes 7 scenarios (S1--S7) which differ in environment, day, and human subject.

To maintain compatibility with previous work, we focus on scenario~S1, in which one ASUS RT-AX86U device generates dummy IEEE~802.11ax (\wifi~6) traffic at a constant rate of 150 frames per second using the frame injection feature of~\cite{gringoliAXCSIEnablingCSI2021}, and a second device acting as \emph{monitor} receives the frames and records the associated \gls{csi} independently for each of its four antennas. Each antenna captures up to 2\,048 \gls{ofdm} subcarriers over a 160\,MHz channel with a subcarrier spacing of 78.125\,kHz, as defined by IEEE~802.11ax. One subject performs each of the following twelve activities in the middle of the room: \emph{Walk}, \emph{Run}, \emph{Jump}, \emph{Sit}, \emph{Empty} (no subject present), \emph{Stand}, \emph{Wave hands}, \emph{Clap}, \emph{Lay down}, \emph{Wipe}, \emph{Squat}, and \emph{Stretch}. Each activity is recorded for 80 seconds, yielding 12\,000 \gls{csi} frames per activity. In this work, we restrict our analysis to a single-antenna configuration; multi-antenna fusion is left as future work.

\subsection{LPCMCI}
PCMCI~\cite{rungeDetectingQuantifyingCausal2019} is a constraint-based causal discovery method for high\-/dimensional, strongly interdependent time series. It comprises two stages. First, a condition selection phase (PC$_1$, a variant of PC~\cite{spirtesAlgorithmFastRecovery1991}) identifies relevant parent sets for each variable through iterative conditional independence (CI) testing. Second, a Momentary Conditional Independence (MCI) test assesses whether pairs of variables are independent given their selected parents. The first stage prunes irrelevant conditions, while the second controls false positives in highly correlated temporal data. PCMCI assumes causal sufficiency, the causal Markov condition, faithfulness, no contemporaneous causal effects, and stationarity.

Latent PCMCI (LPCMCI)~\cite{gerhardusHighrecallCausalDiscovery2020} improves detection power by refining conditioning sets. True causal links may be missed when CI test statistics are weak; LPCMCI restricts conditioning to identified ancestors and augments it with default parent conditions, increasing effect sizes. It further introduces \emph{middle marks}, which encode partial ancestor-descendant information and enable early edge orientation. Edge removal and orientation maintain interpretability throughout the iterative refinement from a fully connected graph to a Partial Ancestral Graph (PAG). Under faithfulness and sufficient samples, \gls{lpcmci} is sound and complete, and accommodates latent confounding while improving recall and false-positive control.

In this work, we apply LPCMCI with partial correlation (ParCorr) as the MCI test. ParCorr evaluates conditional independence via linear partial correlation under Gaussian assumptions~\cite{rungeDetectingQuantifyingCausal2019}. Although our latent variables are binary, ParCorr yields a signed dependence measure whose magnitude reflects effect strength and whose sign encodes excitatory or inhibitory relations. These signed, lagged dependencies are translated into temporally directed logical rules.

\subsection{Linear Temporal Logic}

\gls{ltl}~\cite{pnueliTemporalLogicPrograms1977,baierPrinciplesModelChecking2008} is a widely used formalism for specifying and reasoning about the temporal behavior of reactive systems.\ \gls{ltl} extends classical propositional logic, which evaluates formulas in a single static state, with temporal operators that capture future-oriented behaviors, enabling the specification of constraints and expectations about how a system evolves.

\gls{ltl} formulas describe properties over infinite paths, representing executions of a system modelled as a transition system. We formally introduce \gls{ltl} by defining its syntax, followed by its underlying semantic structure based on transition systems, and finally specifying the satisfaction relation determining when an \gls{ltl} formula holds over an execution path.

\begin{definition}[Syntax of LTL]
    The syntax of \gls{ltl} is defined using the following Backus-Naur form:
    {\small
    \[
        \begin{aligned}
            \phi ::= \quad & \top \mid \bot \mid p \mid (\neg \phi) \mid (\phi \wedge \phi) \mid (\phi \vee \phi) \mid (\phi \rightarrow \phi) \\[1mm]
                           & \mid (\cnext\, \phi) \mid (\ceventually\, \phi) \mid (\cglobally\, \phi) \mid (\phi \cuntil\, \phi),
        \end{aligned}
    \]}
    where \( p \) denotes any propositional atom taken from a designated set of \(\mathit{Atoms}\), $\top$ stands for \textit{true} and $\bot$ stands for \textit{false}. The standard logical connectives (\(\neg, \wedge, \vee, \rightarrow\)) function as in classical propositional logic. The temporal operators allow reasoning over sequences of states:
    \begin{itemize}
        \item \(\cnext\, \phi\) states that \(\phi\) holds in the next state.
        \item \(\ceventually\, \phi\) states that \(\phi\) will hold at some point in the future.
        \item \(\cglobally\, \phi\) states that \(\phi\) holds at all future states.
        \item \(\phi \cuntil \psi\) states that \(\psi\) eventually holds, and until that moment, \(\phi\) holds at every step.
    \end{itemize}
\end{definition}

To interpret \gls{ltl} formulas, we model systems using transition systems, which define the set of states, their transitions, and the atomic propositions that hold in each state.

\begin{definition}[Transition System]
    A transition system provides the semantic framework for interpreting LTL formulas. It is defined as a triple \(\mathcal{M} = (S, \rightarrow, L)\), where:
    \begin{itemize}
        \item \( S \) is a non-empty set of states,
        \item \(\rightarrow \subseteq S \times S\) is a binary transition relation such that for every \( s \in S \), there exists some \( s' \in S \) with \( s \rightarrow s' \),
        \item \( L \colon S \to \mathcal{P}(\mathit{Atoms}) \) is a labelling function assigning to each state the set of atomic propositions true in that state.
    \end{itemize}
\end{definition}

A system's execution is a path: an infinite sequence of states following the transition relation.

\begin{definition}[Path]
    An execution of a system is represented as a \emph{path}, which is an infinite sequence of states \( s_1, s_2, s_3, \ldots \) such that for each \( i \ge 1 \), \( s_i \rightarrow s_{i+1} \). We denote such a path as
    \[
        s_1 \rightarrow s_2 \rightarrow s_3 \rightarrow \cdots.
    \]
\end{definition}

Consider the path \(\pi = s_1 \rightarrow s_2 \rightarrow \cdots\). It represents a possible future system state: initially, the system is in state \(s_1\), then transitions to state \(s_2\), and so on. We write \(\pi^i\) to denote the suffix of \(\pi\) starting at state \(s_i\). For example, \(\pi^3\) represents the execution sequence beginning from \(s_3\), i.e., \(s_3 \rightarrow s_4 \rightarrow \cdots\).

The meaning of an \gls{ltl} formula is defined in terms of its satisfaction over paths. The semantics specify when a formula holds in a given execution trace.

\begin{definition}[Semantics of \gls{ltl}]
    The semantics of LTL determines how formulas are evaluated over paths in a transition system. Given a model \(\mathcal{M} = (S, \rightarrow, L)\) and a path \(\pi = s_1 \rightarrow s_2 \rightarrow s_3 \rightarrow \cdots\), the satisfaction relation \(\vDash\) is defined as follows:
    \begin{enumerate}
        \item \(\pi \vDash \top\) (true).
        \item \(\pi \nvDash \bot\) (false).
        \item \(\pi \vDash p\) if and only if \( p \in L(s_1) \).
        \item \(\pi \vDash \neg \phi\) if and only if \(\pi \nvDash \phi\).
        \item \(\pi \vDash \phi_1 \wedge \phi_2\) if and only if \(\pi \vDash \phi_1\) and \(\pi \vDash \phi_2\).
        \item \(\pi \vDash \phi_1 \vee \phi_2\) if and only if \(\pi \vDash \phi_1\) or \(\pi \vDash \phi_2\).
        \item \(\pi \vDash \phi_1 \rightarrow \phi_2\) if and only if whenever \(\pi \vDash \phi_1\), then \(\pi \vDash \phi_2\).
        \item \(\pi \vDash \cnext\, \phi\) if and only if \(\pi^2 = s_2 \rightarrow s_3 \rightarrow \cdots\) satisfies \(\phi\).
        \item \(\pi \vDash \cglobally\, \phi\) if and only if for all \( i \ge 1 \), \(\pi^i\) satisfies \(\phi\).
        \item \(\pi \vDash \ceventually\, \phi\) if and only if there exists some \( i \ge 1 \) such that \(\pi^i\) satisfies \(\phi\).
        \item \(\pi \vDash \phi \cuntil\, \psi\) if and only if there exists some \( i \ge 1 \) such that \(\pi^i \vDash \psi\) and for all \( j \) with \( 1 \le j < i \), \(\pi^j\) satisfies \(\phi\).
    \end{enumerate}
    Here, \(\pi^i\) denotes the suffix of \(\pi\) starting at the \(i\)th state.
\end{definition}

Since \gls{ltl} formulas are evaluated over paths, we define when a formula holds at a particular state in a transition system.

\begin{definition}[State Satisfaction]
    Since LTL formulas are evaluated over paths, we define how a formula holds at a particular state in a transition system. Given a model \(\mathcal{M} = (S, \rightarrow, L)\), a state \( s \in S \), and an LTL formula \(\phi\), we write
    \[
        \mathcal{M}, s \vDash \phi
    \]
    to indicate that for every path \(\pi\) in \(\mathcal{M}\) that starts at \(s\) (i.e., \(\pi = s \rightarrow s_2 \rightarrow s_3 \rightarrow \cdots\)), we have \(\pi \vDash \phi\).
\end{definition}

\section{Methodology}\label{sec:methodology}
The proposed CHARL-TRE pipeline is structured in three stages:
(i) signal preprocessing and windowing, (ii) categorical latent representation learning via a discrete \gls{vae}, and (iii) causal discovery and deterministic rule extraction in \gls{ltl}. The guiding principle is a strict decoupling between representation learning and symbolic reasoning: the neural component produces a compact discrete latent time series, while the causal-logical component operates exclusively in latent space.

\subsection{CSI Representation and Preprocessing}

Each sample in the considered \gls{csi} dataset (cf.~\cref{ssec:dataset}) is a spectrogram represented as a tensor $\mathbf{X} \in \mathbb{R}^{T \times N_f \times N_a}$, where $T=12\,000$ is the temporal length (eighty seconds of activity with a sampling rate of $150$ frames per second), $N_f=2048$ the number of \wifi subcarriers, and $N_a$ the number of antennas. In this preliminary investigation, we considered a single-antenna configuration ($N_a = 1$). In future work we plan to expand to multi-antenna analysis.

Preprocessing is commonly done in \gls{char} to reduce redundancy and improve robustness while preserving motion-induced channel dynamics~\cite{wangUnderstandingModelingWiFi2015,yousefiSurveyBehaviorRecognition2017}. In particular, we reduced the subcarrier dimensionality by restricting the representation to the first half of the spectrum and downsampling the frequency by retaining one subcarrier every four. Moreover, only the magnitude of the \gls{csi} is retained, discarding phase information to avoid calibration instability. Finally, global min-max normalization is applied over the training set to stabilize optimization.

For representation learning, fixed-length windows of $W_{\mathrm{train}}=75$ samples are extracted, with a stride of $1$. That is, each consecutive window overlaps by $W_{\mathrm{train}}-1$ samples, ensuring that the model learns from all available temporal dynamics. The same windowing strategy is applied to the test set.

\subsection{Categorical Latent Representation Learning}

To obtain a compact and analyzable latent space, we employ a \gls{vae} with categorical latent variables~\cite{kingmaAutoEncodingVariationalBayes2022}. Discrete latent variables are parameterized via the Gumbel-Softmax reparameterization~\cite{jangCategoricalReparameterizationGumbelsoftmax2017,maddisonConcreteDistributionContinuous2017}, enabling gradient-based optimization.

Given an input window $\mathbf{x}$, the encoder defines
\begin{equation}
    q_\phi(\mathbf{z}\mid\mathbf{x}) =
    \prod_{i=1}^{d}
    \mathrm{Cat}\!\left(z_i \mid \boldsymbol{\pi}_i(\mathbf{x})\right),
\end{equation}
where $d$ is the number of latent variables and each $z_i \in \{1,\dots,K\}$. The encoder outputs categorical $\mathrm{Cat}\!\left(z_i \mid \boldsymbol{\pi}_i(\mathbf{x})\right)$ probability vectors $\boldsymbol{\pi}_i(\mathbf{x})$ as $q$.

Sampling is performed using the straight-through Gumbel-Softmax estimator:
\begin{equation}
    \tilde{z}_{ik} =
    \frac{\exp\left((\log \pi_{ik} + g_{ik})/\tau\right)}
    {\sum_{j=1}^{K} \exp\left((\log \pi_{ij} + g_{ij})/\tau\right)},
\end{equation}
with $g_{ik} \sim \mathrm{Gumbel}(0,1)$ and temperature $\tau$. During the forward pass, a hard one-hot projection is applied while gradients are propagated through the relaxed variables.

The decoder reconstructs the \gls{csi} window from the concatenated one-hot latent codes. In the selected configuration, $d=6$ and $K=4$, yielding $dK=24$ binary indicators after flattening.

The VAE architecture utilizes a symmetric encoder-decoder structure optimized for single-antenna CSI magnitude windows. The encoder consists of three 2D convolutional layers with kernel sizes of $(5, 8)$, $(5, 8)$, and $(3, 4)$, and strides of $(5, 8)$, $(5, 8)$, and $(1, 1)$, respectively. Each layer is followed by a \gls{gelu} activation function to capture non-linear temporal and frequency-domain features. The resulting feature map is flattened and processed by two linear heads to parameterize the latent distribution. The decoder mirrors this architecture, employing a linear layer followed by three transpose convolutional layers to reconstruct the input CSI window from the latent representation.

\subsection{Objective Function and Information Capacity Control}

Training minimizes
\begin{equation}
    \mathcal{L} =
    \mathcal{L}_{\mathrm{rec}} +
    \lambda_{\mathrm{KL}} \mathcal{L}_{\mathrm{KL}},
\end{equation}
where $\mathcal{L}_{\mathrm{rec}}$ is the reconstruction term and $\mathcal{L}_{\mathrm{KL}}$ enforces regularization towards a uniform categorical prior, weighted by $\lambda_{\mathrm{KL}}$.

The prior is uniformly factorized as $p(z_i)=\mathrm{Cat}(1/K)$. To mitigate posterior collapse and regulate the information content of the latent representation, we adopt a capacity-controlled formulation inspired by $\beta$-\gls{vae}~\cite{higginsBetaVAELearningBasic2017}:
\begin{equation}
    \mathcal{L}_{\mathrm{KL}} =
    \frac{1}{B}
    \sum_{b=1}^{B}
    \max\!\left(
    0,
    \sum_{i=1}^{d}
    \mathrm{KL}\!\left[
            q_\phi(z_i\mid x_b)\,\|\,p(z_i)
            \right]
    - C
    \right),
\end{equation}
where $C$ is a target capacity and $B$ the batch size.

Three coordinated schedules are employed during each training epoch: monotonic reduction of $\tau$, gradual increase of $\lambda_{\mathrm{KL}}$, and linear ramp-up of $C$. This ensures progressive discretization and controlled information flow into the latent variables.

After training, each input window is mapped through an encoder pass to a hard categorical latent code, which is then the subject of the following causal analysis.

\section{Identifying Temporal Patterns}\label{sec:causal}
There is a need to establish the structural bridge between unsupervised discrete compression of \gls{csi} and deterministic symbolic recognition. Instead of attaching an additional neural classifier to the learned latent representation, we analyze its temporal organization explicitly and transform statistically inferred dependencies into compact \gls{ltl} specifications. The central claim is that a categorical latent space learned without supervision encodes class-specific temporal structure that can be recovered, formalized, and used as a fully deterministic classifier. The objective is not to label latent variables, but to determine whether their temporal interactions are structured and discriminative across activities.

\subsection{Class-Conditional Temporal Structure Estimation}

Let $\mathbf{Z}_t$ denote the binary one-hot encoding of the categorical latent variables at time $t$. For each activity $a \in \mathcal{A}$, CHARL-TRE collects the latent trajectories corresponding to windows labelled as $a$ and performs causal discovery separately for each class. This class-conditional approach yields activity-specific lagged dependency graphs over latent propositions.

For each ordered pair of source \(s\) and destination \(d\) latent propositions $(X_s,X_d)$ (not necessarily belonging to the same time vector) and lag $\tau$, \gls{lpcmci} computes a momentary conditional independence statistic conditioned on an adaptively selected set of parents. We considered time lags $\tau \in [1,5]$ to capture short-term temporal dependencies while maintaining computational tractability.

The sign of this conditional dependence statistic is essential. A positive value indicates that, under the chosen conditioning set, activation of $X_s(t)$ is associated with an increased conditional expectation of $X_d(t+\tau)$. A negative value indicates a suppressive or antagonistic conditional association. We interpret these signed conditional effects as oriented influences in latent state space. Consequently, the polarity of the discovered dependence determines whether the corresponding temporal rule encodes activation or exclusion. The logical translation is therefore grounded directly in the signed conditional dependence returned by \gls{lpcmci}, rather than imposed externally.

\subsection{From Signed Dependencies to Temporal Logic}

Each retained lagged dependency from $X_s(t)$ to $X_d(t+\tau)$ is mapped deterministically to an \gls{ltl} formula. If the conditional dependence is positive, we construct
\begin{equation}
    \Box\!\left(X_s \rightarrow \bigcirc^{\tau} X_d\right),
\end{equation}
and if the conditional dependence is negative, we construct
\begin{equation}
    \Box\!\left(X_s \rightarrow \bigcirc^{\tau} \neg X_d\right).
\end{equation}
Here $\Box$ denotes the global operator and $\bigcirc^{\tau}$ the $\tau$-fold next operator. Positive conditional effects induce temporal activation rules, while negative conditional effects induce temporal exclusion constraints. This translation preserves temporal direction, lag, and the orientation of the estimated conditional influence.

For each activity $a$, this procedure yields a pool of candidate temporal implications derived directly from its class-conditional causal graph.

\subsection{Two-Level Search for the Best Deterministic Configuration}

The final symbolic classifier is obtained through a structured two-level search.

\subsubsection{Rule Selection Within a Fixed Configuration}

A configuration is defined by segmentation parameters (window length and hop), lag horizon, graph filtering thresholds, maximum rule budget per activity, and scoring parameters. For a fixed configuration, rule selection proceeds independently for each activity.

For each candidate rule $\phi_{a,i}$, we compute the number of antecedent occurrences $n_{a,i}$ and the number of satisfied consequents $h_{a,i}$ on training data, with lag $\tau_{a,i}$:
\begin{align}
    n_{a,i}(\mathbf{x}) & = \sum_{t=0}^{T-\tau_{a,i}} \mathbbm{1}[\text{antecedent}_{a,i}(t)], \label{eq:antecedents}                                              \\
    h_{a,i}(\mathbf{x}) & = \sum_{t=0}^{T-\tau_{a,i}} \mathbbm{1}[\text{antecedent}_{a,i}(t) \wedge \text{consequent}_{a,i}(t+\tau_{a,i})]. \label{eq:consequents}
\end{align}
With Laplace smoothing, we compute
\begin{equation}
    \hat p_{a,i} = \frac{h_{a,i}+1}{n_{a,i}+2},
\end{equation}
along with its analogous estimate $\hat p_{\neg a,i}$ evaluated over all other activities. A discriminative margin
\begin{equation}\label{eq:margin}
    \Delta_{a,i} = \hat p_{a,i} - \hat p_{\neg a,i}
\end{equation}
measures how much more reliably the rule holds in its own class than elsewhere.

Rules with positive margin are retained and ranked according to:
\[
    w_r=\omega_r \max(0,\Delta_r)\log(1+n_r),
\]
where \(\omega_r\) denotes the causal edge strength, i.e.\ the absolute value of the conditional dependence statistic returned by \gls{lpcmci} for the corresponding lagged edge, \(n_r\) the empirical support of the rule, computed as the number of instances in the dataset that satisfy the rule's antecedent conditions, and \(\Delta_r=\hat p_r^{(a)}-\hat p_r^{(\neg a)}\) the discriminative margin between the target activity \(a\) and all non-target activities. The top-ranked rules up to the allowed budget define the rule set $\Phi_a$ for that activity under the current configuration.

\subsubsection{Deterministic Classifier Construction}

Given the selected rule sets ${\{\Phi_a\}}_{a\in\mathcal{A}}$, a deterministic classifier is constructed. For an evaluation segment $\mathbf{x}$ and rule $\phi_{a,i}$ with lag $\tau_{a,i}$, we identify $n_{a,i}(\mathbf{x})$ and $h_{a,i}(\mathbf{x})$ with the formulas in \Cref{eq:antecedents,eq:consequents}, respectively. Each rule contributes a log-likelihood ratio
\begin{equation}
    \mathrm{LLR}_{a,i}(\mathbf{x}) =
    h_{a,i}\log\frac{\hat p_{a,i}}{\hat p_{\neg a,i}}
    +
    (n_{a,i}-h_{a,i})
    \log\frac{1-\hat p_{a,i}}{1-\hat p_{\neg a,i}}.
\end{equation}
The class score is
\begin{equation}
    S_a^{\mathrm{rule}}(\mathbf{x}) =
    \sum_{i=1}^{N_a} \widetilde w_{a,i}\,\mathrm{LLR}_{a,i}(\mathbf{x}),
\end{equation} where $\widetilde w_{a,i}$ are normalized weights for each rule, typically derived from their ranking scores. The predicted class is the one with the highest rule-based score:
\begin{equation}
    \hat a = \arg\max_{a\in\mathcal{A}} S_a^{\mathrm{rule}}(\mathbf{x}),
\end{equation}
with an optional prototype fallback for classes without selected rules.

\subsubsection{Search Across Configurations}

The best-performing configuration adopts a chronological train/test split of \(0.7/0.3\) per activity, segment length \(80\), hop size \(10\), and classifier stride \(75\). Rule extraction is performed on the directed lagged edges of the path graph with a pruning threshold of \(\theta_p=0.0\), i.e.\ no pruning by edge strength is applied at the rule-search stage. This choice maximizes the recall of potentially discriminative edges at the cost of admitting weaker associations. In practice, activities with subtle or low-support temporal patterns, such as \emph{Waving} and \emph{Clap}, would lose candidate rules under any positive \(\theta_p\). A separate graph filter pruning threshold \(\theta_f=0.1\) is used exclusively for filtered-graph summaries and visualization; it has no effect on classification.

Classification operates purely on the selected symbolic rules (core-fraction \(=1.0\)). A prototype-based fallback term with weight \(1.0\) is activated only for activities for which no symbolic rules are selected. Concretely, for such an activity \(a\) with a dataset of training segments \(\mathcal{D}_a\), a prototype latent vector
\[
    \mu_a = \frac{1}{|\mathcal{D}_a|}\sum_{x\in\mathcal{D}_a}\bar{x}
\]
is computed from the mean latent activations \(\bar{x}\) of its training segments. For a test segment \(x\), the fallback score is
\[
    P_a(x) = -\frac{1}{M}\sum_{m=1}^{M}\left|\bar{x}_m-\mu_{a,m}\right|,
\]
where \(M\) is the number of latent propositions and \(\mu_{a,m}\) is the mean activation of latent variable \(Z_m\) for activity \(a\). This term ensures that activities without symbolic support remain classifiable, whilst preserving a fully rule-based decision mechanism whenever symbolic rules are available. In the reported run, this mechanism is effectively used only for \textit{Empty}.

The best configuration is therefore defined as the one achieving the highest held-out deterministic accuracy subject to minimal support constraints. Conceptually, the procedure can be summarized as follows: generate causal lagged implications in the latent space, retain those that most separate each activity from the others, and tune segmentation and scoring parameters until deterministic majority voting on unseen data is optimized.

Let $z_{d,c}$ denote the atomic proposition that latent variable $Z_d$ is in category $c$ at the current time step. The full deterministic rule base obtained in the best configuration is given below.
\\
\paragraph{Walk ($N=15$)}
{\scriptsize
    \begin{flalign*}
                                                              & \Box(z_{0,3}\!\rightarrow\!\bigcirc^{2}z_{1,2}),\;
        \Box(z_{0,3}\!\rightarrow\!\bigcirc^{2}z_{0,3}),\;
        \Box(z_{1,2}\!\rightarrow\!\bigcirc^{2}z_{1,2}),      &                                                         & \\
                                                              & \Box(z_{1,2}\!\rightarrow\!\bigcirc^{2}z_{0,3}),\;
        \Box(z_{1,2}\!\rightarrow\!\bigcirc^{2}\neg z_{1,3}),\;
        \Box(z_{0,3}\!\rightarrow\!\bigcirc^{2}\neg z_{0,2}), &                                                         & \\
                                                              & \Box(z_{0,3}\!\rightarrow\!\bigcirc^{2}\neg z_{0,0}),\;
        \Box(z_{0,3}\!\rightarrow\!\bigcirc^{2}\neg z_{1,1}),\;
        \Box(z_{0,3}\!\rightarrow\!\bigcirc^{4}\neg z_{0,2}), &                                                         & \\
                                                              & \Box(z_{5,1}\!\rightarrow\!\bigcirc^{1}z_{3,1}),\;
        \Box(z_{1,2}\!\rightarrow\!\bigcirc^{2}z_{5,2}),\;
        \Box(z_{5,3}\!\rightarrow\!\bigcirc^{2}z_{1,3}),      &                                                         & \\
                                                              & \Box(z_{5,3}\!\rightarrow\!\bigcirc^{2}z_{4,2}),\;
        \Box(z_{5,3}\!\rightarrow\!\bigcirc^{2}z_{5,3}),\;
        \Box(z_{3,0}\!\rightarrow\!\bigcirc^{2}z_{0,1})
    \end{flalign*}
}

\paragraph{Run ($N=10$)}
{\scriptsize
    \begin{flalign*}
                                                                & \Box(z_{0,3}\!\rightarrow\!\bigcirc^{2}z_{1,2}),\;
        \Box(z_{1,2}\!\rightarrow\!\bigcirc^{2}z_{1,2}),\;
        \Box(z_{1,2}\!\rightarrow\!\bigcirc^{2}z_{0,3}),        &                                                         & \\
                                                                & \Box(z_{0,3}\!\rightarrow\!\bigcirc^{2}z_{0,3}),\;
        \Box(z_{0,3}\!\rightarrow\!\bigcirc^{2}\neg z_{0,2}),\;
        \Box(z_{1,2}\!\rightarrow\!\bigcirc^{2}\neg z_{1,3}),   &                                                         & \\
                                                                & \Box(z_{1,2}\!\rightarrow\!\bigcirc^{2}\neg z_{0,2}),\;
        \Box(z_{1,2}\!\rightarrow\!\bigcirc^{2}\neg z_{0,1}),\;
        \Box(z_{0,3}\!\rightarrow\!\bigcirc^{2}\neg z_{1,1}),\; &                                                         & \\
                                                                & \Box(z_{1,2}\!\rightarrow\!\bigcirc^{4}\neg z_{5,0})
    \end{flalign*}
}

\paragraph{Jump ($N=12$)}
{\scriptsize
    \begin{flalign*}
                                                              & \Box(z_{0,2}\!\rightarrow\!\bigcirc^{4}z_{1,3}),\;
        \Box(z_{5,3}\!\rightarrow\!\bigcirc^{2}z_{1,3}),\;
        \Box(z_{1,2}\!\rightarrow\!\bigcirc^{2}z_{0,3}),                                                                  \\
                                                              & \Box(z_{1,2}\!\rightarrow\!\bigcirc^{2}\neg z_{1,3}),\;
        \Box(z_{4,3}\!\rightarrow\!\bigcirc^{5}z_{2,0}),\;
        \Box(z_{0,0}\!\rightarrow\!\bigcirc^{4}z_{1,1}),                                                                  \\
                                                              & \Box(z_{3,1}\!\rightarrow\!\bigcirc^{2}\neg z_{5,0}),\;
        \Box(z_{1,3}\!\rightarrow\!\bigcirc^{4}z_{0,1}),\;
        \Box(z_{1,2}\!\rightarrow\!\bigcirc^{2}\neg z_{0,0}), &                                                         & \\
                                                              & \Box(z_{0,3}\!\rightarrow\!\bigcirc^{1}z_{3,2}),\;
        \Box(z_{1,2}\!\rightarrow\!\bigcirc^{5}\neg z_{0,0}),\;
        \Box(z_{1,2}\!\rightarrow\!\bigcirc^{5}z_{0,3})       &                                                         &
    \end{flalign*}
}

\paragraph{Sit ($N=14$)}
{\scriptsize
    \begin{flalign*}
                                                         & \Box(z_{5,3}\!\rightarrow\!\bigcirc^{2}z_{5,3}),\;
        \Box(z_{5,3}\!\rightarrow\!\bigcirc^{4}z_{5,3}),\;
        \Box(z_{5,3}\!\rightarrow\!\bigcirc^{4}z_{4,2}),                                                             \\
                                                         & \Box(z_{5,3}\!\rightarrow\!\bigcirc^{2}z_{4,2}),\;
        \Box(z_{5,3}\!\rightarrow\!\bigcirc^{4}z_{0,0}),\;
        \Box(z_{1,3}\!\rightarrow\!\bigcirc^{5}\neg z_{3,3}),                                                        \\
                                                         & \Box(z_{5,3}\!\rightarrow\!\bigcirc^{4}\neg z_{1,2}),\;
        \Box(z_{3,1}\!\rightarrow\!\bigcirc^{1}z_{0,2}),\;
        \Box(z_{4,1}\!\rightarrow\!\bigcirc^{2}z_{5,1}), &                                                         & \\
                                                         & \Box(z_{3,2}\!\rightarrow\!\bigcirc^{4}\neg z_{5,2}),\;
        \Box(z_{5,2}\!\rightarrow\!\bigcirc^{3}z_{1,2}),\;
        \Box(z_{5,2}\!\rightarrow\!\bigcirc^{2}z_{5,2}), &                                                         & \\
                                                         & \Box(z_{5,2}\!\rightarrow\!\bigcirc^{3}z_{0,3}),\;
        \Box(z_{3,1}\!\rightarrow\!\bigcirc^{1}z_{3,1})
    \end{flalign*}
}

\paragraph{Empty ($N=0$)}
{\scriptsize
    \begin{flalign*}
         & \Phi_{\mathrm{Empty}} = \varnothing. &  &
    \end{flalign*}
}

\paragraph{Stand ($N=11$)}
{\scriptsize
    \begin{flalign*}
                                                         & \Box(z_{5,3}\!\rightarrow\!\bigcirc^{4}\neg z_{5,0}),\;
        \Box(z_{5,3}\!\rightarrow\!\bigcirc^{2}\neg z_{5,0}),\;
        \Box(z_{5,3}\!\rightarrow\!\bigcirc^{2}z_{4,2}), &                                                         & \\
                                                         & \Box(z_{1,2}\!\rightarrow\!\bigcirc^{2}z_{1,2}),\;
        \Box(z_{1,2}\!\rightarrow\!\bigcirc^{2}z_{5,2}),\;
        \Box(z_{5,2}\!\rightarrow\!\bigcirc^{4}z_{4,3}), &                                                         & \\
                                                         & \Box(z_{1,2}\!\rightarrow\!\bigcirc^{4}\neg z_{0,3}),\;
        \Box(z_{1,1}\!\rightarrow\!\bigcirc^{1}z_{3,2}),\;
        \Box(z_{5,2}\!\rightarrow\!\bigcirc^{4}z_{1,2}), &                                                         & \\
                                                         & \Box(z_{1,2}\!\rightarrow\!\bigcirc^{2}z_{0,3}),\;
        \Box(z_{1,2}\!\rightarrow\!\bigcirc^{4}\neg z_{1,2})
    \end{flalign*}
}

\paragraph{Waving ($N=4$)}
{\scriptsize
    \begin{flalign*}
                                                         & \Box(z_{0,3}\!\rightarrow\!\bigcirc^{2}z_{0,3}),\;
        \Box(z_{0,3}\!\rightarrow\!\bigcirc^{2}\neg z_{0,0}),\;
        \Box(z_{0,1}\!\rightarrow\!\bigcirc^{2}z_{0,1}), &                                                    & \\
                                                         & \Box(z_{1,3}\!\rightarrow\!\bigcirc^{2}z_{5,1})
    \end{flalign*}
}

\paragraph{Clap ($N=2$)}
{\scriptsize
    \begin{flalign*}
                                                        & \Box(z_{3,3}\!\rightarrow\!\bigcirc^{1}z_{2,3}),\;
        \Box(z_{3,1}\!\rightarrow\!\bigcirc^{3}z_{1,2}) &                                                    &
    \end{flalign*}
}

\paragraph{Lay Down ($N=2$)}
{\scriptsize
    \begin{flalign*}
                                                             & \Box(z_{5,3}\!\rightarrow\!\bigcirc^{5}z_{5,3}),\;
        \Box(z_{5,0}\!\rightarrow\!\bigcirc^{3}\neg z_{4,3}) &                                                    &
    \end{flalign*}
}

\paragraph{Wipe ($N=12$)}
{\scriptsize
    \begin{flalign*}
                                                         & \Box(z_{0,3}\!\rightarrow\!\bigcirc^{2}z_{0,3}),\;
        \Box(z_{1,2}\!\rightarrow\!\bigcirc^{2}z_{0,3}),\;
        \Box(z_{1,2}\!\rightarrow\!\bigcirc^{2}z_{1,2}), &                                                         & \\
                                                         & \Box(z_{0,3}\!\rightarrow\!\bigcirc^{2}\neg z_{0,2}),\;
        \Box(z_{1,2}\!\rightarrow\!\bigcirc^{2}\neg z_{0,0}),\;
        \Box(z_{1,2}\!\rightarrow\!\bigcirc^{4}z_{1,2}), &                                                         & \\
                                                         & \Box(z_{5,2}\!\rightarrow\!\bigcirc^{2}z_{1,2}),\;
        \Box(z_{5,1}\!\rightarrow\!\bigcirc^{4}z_{0,0}),\;
        \Box(z_{1,3}\!\rightarrow\!\bigcirc^{2}z_{4,0}), &                                                         & \\
                                                         & \Box(z_{5,3}\!\rightarrow\!\bigcirc^{5}z_{5,3}),\;
        \Box(z_{5,2}\!\rightarrow\!\bigcirc^{2}z_{3,2}),\;
        \Box(z_{1,1}\!\rightarrow\!\bigcirc^{2}z_{5,1})  &                                                         &
    \end{flalign*}
}

\paragraph{Squat ($N=9$)}
{\scriptsize
    \begin{flalign*}
                                                              & \Box(z_{5,3}\!\rightarrow\!\bigcirc^{2}z_{5,3}),\;
        \Box(z_{4,2}\!\rightarrow\!\bigcirc^{2}z_{5,3}),\;
        \Box(z_{5,3}\!\rightarrow\!\bigcirc^{2}\neg z_{5,0}), &                                                    & \\
                                                              & \Box(z_{5,0}\!\rightarrow\!\bigcirc^{3}z_{0,0}),\;
        \Box(z_{0,0}\!\rightarrow\!\bigcirc^{2}\neg z_{5,0}),
        \Box(z_{5,0}\!\rightarrow\!\bigcirc^{3}\neg z_{0,2}), &                                                    & \\
                                                              & \Box(z_{1,2}\!\rightarrow\!\bigcirc^{1}z_{1,2}),\;
        \Box(z_{4,3}\!\rightarrow\!\bigcirc^{2}z_{5,2}),\;
        \Box(z_{3,1}\!\rightarrow\!\bigcirc^{5}z_{3,3})       &                                                    &
    \end{flalign*}
}

\paragraph{Stretch ($N=4$)}
{\scriptsize
    \begin{flalign*}
                                                         & \Box(z_{0,2}\!\rightarrow\!\bigcirc^{1}z_{5,1}),\;
        \Box(z_{5,3}\!\rightarrow\!\bigcirc^{2}z_{5,3}),\;
        \Box(z_{5,3}\!\rightarrow\!\bigcirc^{4}z_{5,3}), &                                                    &     \\
                                                         & \Box(z_{0,3}\!\rightarrow\!\bigcirc^{2}z_{0,3})    &   &
    \end{flalign*}
}

\section{Preliminary Experimental Results}\label{sec:results}
The results of the pipeline evaluation indicate that statistically inferred temporal structure in an unsupervised discrete latent space can sustain competitive deterministic classification, with accuracy exceeding 70\% despite the absence of end-to-end supervision or probabilistic calibration.

\begin{figure}[t]
    \centering
    \includegraphics[width=.8\columnwidth]{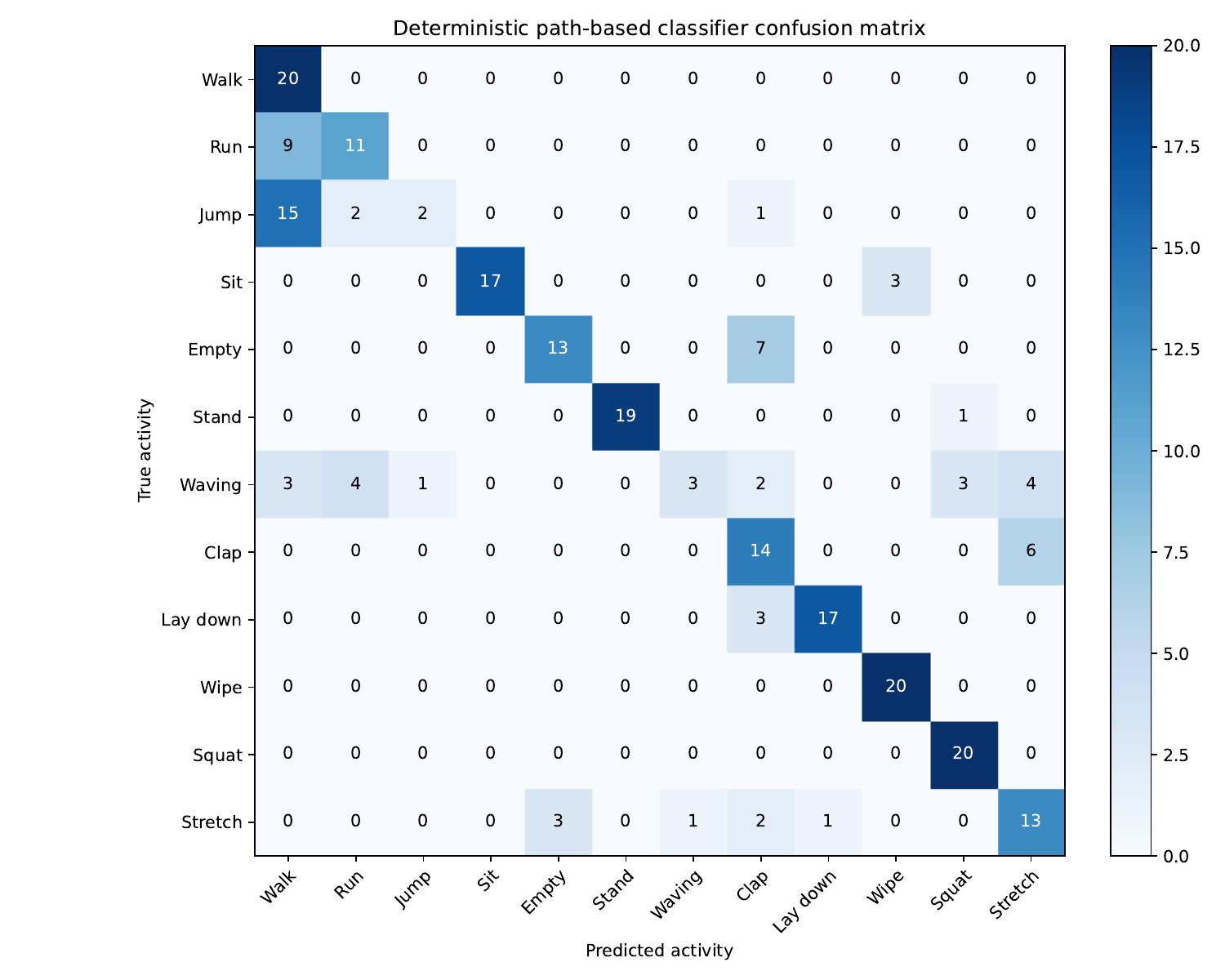}
    \caption{Confusion matrix of the best deterministic symbolic classifier (ParCorr-based causal graphs, fully decoupled rule and classifier tuning).}\label{fig:deterministic-confusion-best}
\end{figure}

\paragraph{Activity complexity and classification errors}
The confusion matrix (\Cref{fig:deterministic-confusion-best}) reveals a clear correlation between activity complexity and recognition performance. Quasi-stationary activities, such as \emph{Stand}, \emph{Wipe}, \emph{Squat}, \emph{Sit}, and \emph{Lay down}, are recognized with high recall and precision, suggesting that their latent trajectories exhibit stable and discriminative temporal dependencies that are well captured by pairwise lagged rules. In contrast, highly dynamic activities exhibit the opposite trend: \emph{Jump} (15 of 20 segments misclassified as \emph{Walk}) and \emph{Waving} ($N=4$ rules) suffer from non-stationary latent trajectories that yield few high-margin edges, and their residual evidence is absorbed by the more generic \emph{Walk} rule set. The \emph{Empty}--\emph{Clap} confusion (7 of 20) reflects the prototype fallback: without symbolic rules, \emph{Empty} is classified via mean latent activations, which overlap with the sparse signature of \emph{Clap}. More broadly, activities with low rule counts, such as  \emph{Clap} ($N=2$), \emph{Lay down} ($N=2$), or \emph{Waving} ($N=4$), are the most error-prone, confirming that rule budget is a reliable proxy for classification difficulty.

The origin of these behaviors can be clarified by the strict decoupling between representation learning, causal discovery, and rule induction. Because causal graphs are estimated once and subsequently reused, performance variations arise from rule selection and aggregation rather than from stochastic graph re-estimation. This separation improves reproducibility and enables systematic exploration of symbolic hyperparameters with minimal computational overhead.

Moreover, the symbolic formulation enables principled multi-antenna fusion at the rule level. Because classification is expressed through explicit \gls{ltl} formulae over discrete latent variables, rule sets extracted independently from different antennas can be combined through logical conjunction, disjunction, or voting schemes using the same encoder. Empirical observations indicate that this modular aggregation can exploit spatial diversity and reduce ambiguity in cases where single-antenna evidence is weak, while preserving deterministic behavior and interpretability. This suggests a structured avenue for extending the framework to multi-receiver settings without altering the underlying representation learning stage.

\paragraph{Interpretability}
A key advantage of \gls{charl-tre} is that every classification decision is fully traceable. Each rule in $\Phi_a$ is a concrete \gls{ltl} formula over named latent propositions $z_{d,c}$, with an associated lag $\tau$, empirical support $n_{a,i}$, and discriminative margin $\Delta_{a,i}$, all of which are directly inspectable. For example, the rule $\square(z_{0,3} \rightarrow \bigcirc^2 z_{1,2})$ extracted for \emph{Walk} states that whenever latent variable $Z_0$ is in category 3, variable $Z_1$ will be in category 2 two time steps later: a temporally directed regularity that recurs across \emph{Walk} segments but not others. The primary limitation is that the latent propositions $z_{d,c}$ themselves carry no predefined semantics; linking them to physical \gls{csi} phenomena remains an open problem.

\paragraph{Scalability and computational considerations}
The scalability of the proposed pipeline is addressed through the modularity of the \gls{vae} architecture and the dimensionality reduction achieved via strided convolutions. By employing large strides, the encoder significantly reduces the high-dimensional \gls{csi} input early in the processing chain. The VAE training scales also efficiently with large datasets via stochastic gradient descent~\cite{kingmaAutoEncodingVariationalBayes2022}. The downstream causal discovery process remains the primary computational bottleneck as it depends exponentially on the number of extracted latent variables.

\paragraph{Limitations}
Compression into a fixed-capacity categorical latent space inevitably discards information, potentially limiting separability for activities with subtle or rapidly changing signatures. ParCorr-based \gls{lpcmci} relies on a linear conditional dependence statistic, which may fail to detect non-linear interactions among latent variables. Temporal dependencies are restricted to lags \(\tau \leq 5\), constraining the model's ability to capture longer-range structure. Moreover, the evaluation set comprises 240 segments, limiting statistical power in edge estimation and increasing sensitivity to sampling variability. The relatively permissive rule thresholds adopted in the best configuration further suggest that broader inclusion of weak dependencies was beneficial, but this may also amplify noise in low-support regimes.

\paragraph{Overall results}
The findings demonstrate that a fully deterministic \gls{ltl} classifier derived from statistically inferred latent temporal relations can achieve meaningful performance while preserving symbolic transparency. The approach offers a controlled trade-off between interpretability and accuracy, although improvements are likely to require richer dependency tests, extended temporal horizons, or more expressive latent structures.

\section{Conclusion}\label{sec:conclusions}
\gls{charl-tre} is a fully automatic pipeline for \wifi \gls{csi}-based\gls{har} that reconciles raw signal processing with causal interpretability and deterministic symbolic classification. A categorical variational autoencoder compresses \gls{csi} windows into a compact discrete latent space; causal discovery is then performed exclusively on the resulting one-hot latent trajectories, and statistically supported temporal dependencies are translated into \gls{ltl} rules that form a fully deterministic classifier. Representation learning, causal analysis, and rule induction are strictly decoupled, so rules can be inspected, pruned, or modified without retraining the encoder, and antenna-specific rule sets can be combined through logical aggregation to support principled multi-antenna fusion.

Experimentally, \gls{charl-tre} achieves competitive accuracy despite relying entirely on \gls{ltl} rule evaluation rather than a learned discriminative head. Notably, although the latent variables carry no predefined semantics, their temporal interaction structure proves stable enough to sustain consistent rule extraction — suggesting that semantic interpretability at the level of individual latent dimensions is not a prerequisite for causal and symbolic reasoning.

The approach has three main limitations. First, discrete compression inevitably discards information relative to the original \gls{csi} signal, potentially limiting separability for activities with subtle or rapidly changing signatures. Second, causal discovery relies on linear conditional independence tests restricted to short temporal lags ($\tau \leq 5$), which may miss non-linear or longer-range dependencies. Third, the evaluation set of 240 segments constrains the statistical power of graph estimation. Future work will therefore explore non-linear causal tests, extended temporal horizons, larger evaluation sets, and robustness under domain shift.

Overall, these results establish deterministic \gls{ltl} classifiers grounded in unsupervised discrete latent representations as a viable and principled alternative to end-to-end black-box models for wireless \gls{har}, offering explicit causal structure and symbolic controllability without sacrificing practical performance.

\section*{Acknowledgments}
The work was partially supported by the European Office of Aerospace Research \& Development under award numbers FA8655\--22\--1\--7017 and FA8655\--25\--1\--7067, and by the U.S. Army DEVCOM Army Research Laboratory (ARL) under Cooperative Agreement \#W911NF2220243. Any opinions, findings, and conclusions or recommendations expressed in this material are those of the author(s) and do not necessarily reflect the views of the authors or of the United States government.

The authors used Claude Sonnet 4.6 to improve readability and language. They reviewed and edited the content as needed, and they take full responsibility for the publication's content.

\bibliographystyle{IEEEtran}
\bibliography{references}

\end{document}